\documentclass[graybox]{svmult}

\usepackage{type1cm}        
\usepackage{makeidx}         
\usepackage{graphicx}        
\usepackage{multicol}        
\usepackage[bottom]{footmisc}

\usepackage{newtxtext}       %
\usepackage[varvw]{newtxmath}       

\makeindex             

\DeclareMathOperator{\atantwo}{atan2}
\DeclareMathOperator{\acos}{acos}
\DeclareMathOperator{\asinh}{asinh}
\DeclareMathOperator*{\argmin}{\arg\!\min}
\DeclareMathOperator{\Tr}{Tr}
\newcommand{\norm}[1]{\left\lVert#1\right\rVert}

\begin{document}

\title*{Preliminary Analysis and Simulation of a Compact Variable Stiffness Wrist}
\titlerunning{Preliminary Analysis and Simulation of a Compact Variable Stiffness Wrist} 
\author{Giuseppe Milazzo, Manuel G. Catalano, Antonio Bicchi, and Giorgio Grioli}
\authorrunning{Giuseppe Milazzo, Manuel G. Catalano, Antonio Bicchi, and Giorgio Grioli}
\institute{Giuseppe Milazzo \at Soft Robotics for Human Cooperation and Rehabilitation, Istituto Italiano di Tecnologia,  Via S. Quirico, 19d, 16163, Genova (GE), \email{giuseppe.milazzo@iit.it}
\and Manuel Giuseppe Catalano \at Soft Robotics for Human Cooperation and Rehabilitation, Istituto Italiano di Tecnologia,  Via S. Quirico, 19d, 16163, Genova (GE)
\and Antonio Bicchi \at Soft Robotics for Human Cooperation and Rehabilitation, Istituto Italiano di Tecnologia,  Via S. Quirico, 19d, 16163, Genova (GE). Centro di Ricerca E. Piaggio \& Dipartimento di Ingegneria dell'Informazione, Università di Pisa, Via Diotisalvi, 1, 56122, Pisa (PI)
\and Giorgio Grioli \at Soft Robotics for Human Cooperation and Rehabilitation, Istituto Italiano di Tecnologia,  Via S. Quirico, 19d, 16163, Genova (GE). Centro di Ricerca E. Piaggio \& Dipartimento di Ingegneria dell'Informazione, Università di Pisa, Via Diotisalvi, 1, 56122, Pisa (PI)}
\maketitle
\abstract*{Variable Stiffness Actuators prove invaluable for robotics applications in unstructured environments, fostering safe interactions and enhancing task adaptability. Nevertheless, their mechanical design inevitably results in larger and heavier structures compared to classical rigid actuators. \\
\indent This paper introduces a novel 3 Degrees of Freedom (DoFs) parallel wrist that achieves variable stiffness through redundant elastic actuation. Leveraging its parallel architecture, the device employs only four motors, rendering it compact and lightweight. This characteristic makes it particularly well-suited for applications in prosthetics or humanoid robotics.\\
\indent The manuscript delves into the theoretical model of the device and proposes a sophisticated control strategy for independent regulation of joint position and stiffness.
Furthermore, it validates the proposed controller through simulation, utilizing a comprehensive analysis of the system dynamics.
The reported results affirm the ability of the device to achieve high accuracy and disturbance rejection in rigid configurations while minimizing interaction forces with its compliant behavior.}
\abstract{Variable Stiffness Actuators prove invaluable for robotics applications in unstructured environments, fostering safe interactions and enhancing task adaptability. Nevertheless, their mechanical design inevitably results in larger and heavier structures compared to classical rigid actuators. \\
\indent This paper introduces a novel 3 Degrees of Freedom (DoFs) parallel wrist that achieves variable stiffness through redundant elastic actuation. Leveraging its parallel architecture, the device employs only four motors, rendering it compact and lightweight. This characteristic makes it particularly well-suited for applications in prosthetics or humanoid robotics.\\
\indent The manuscript delves into the theoretical model of the device and proposes a sophisticated control strategy for independent regulation of joint position and stiffness.
Furthermore, it validates the proposed controller through simulation, utilizing a comprehensive analysis of the system dynamics.
The reported results affirm the ability of the device to achieve high accuracy and disturbance rejection in rigid configurations while minimizing interaction forces with its compliant behavior.}


\section{Introduction}
\label{sec:Introduction}

Wrist articulation plays a pivotal role in facilitating grasping tasks by enabling manipulators to effortlessly reposition the end-effector to reach optimal grasping poses.
Humans excel in tooling tasks, effortlessly adapting to mechanical and geometrical constraints through the inherent compliance of their wrist articulation \cite{Phan2020Tooling}.
However, replicating these tasks with conventional industrial robots solely through software control remains an arduous challenge. 
Instead, utilizing Variable Stiffness (VS) joints allows robots to exploit environmental constraints, capitalizing on inherent compliance, while maintaining a stiff configuration to resist external perturbations and lift heavy loads, mirroring human capabilities \cite{Osu2004Optimal,Borzelli2018Muscles}.

Variable Stiffness Actuators (VSAs) enhance task adaptability and ensure safe interactions by incorporating reconfigurable compliant elements into the mechanical design of the device \cite{Wolf2016VSA}. Nevertheless, operating an $n$-Degrees of Freedom (DoFs) Variable Stiffness (VS) joint typically demands $2n$ actuators, resulting in augmented device size and weight, along with heightened complexity in mechanical design \cite{Grioli2015User}.
In \cite{Lemerle2021Wrist}, the authors introduce a novel VS 2 DoFs joint that leverages its parallel architecture to achieve VS using only 3 motors. Building upon this concept, the authors in \cite{Milazzo2023Wrist} utilized the aforementioned VS joint to craft a 3 DoFs VS wrist that, thanks to its compact design, holds promise for applications in prosthetics or humanoid robotics.

This study extends the theoretical model of the device outlined in \cite{Milazzo2023Wrist}, offering comprehensive insights into its kinematics and dynamics. Subsequently, the manuscript validates the proposed device concept and control strategy through simulations. Section 2 elucidates the working principles of the system, while Section 3 details the theoretical model. The simulation results are reported in Section 4, and conclusions are drawn in Section 5.

\begin{figure}[b]
\sidecaption[t]
    \includegraphics[width = 7 cm]{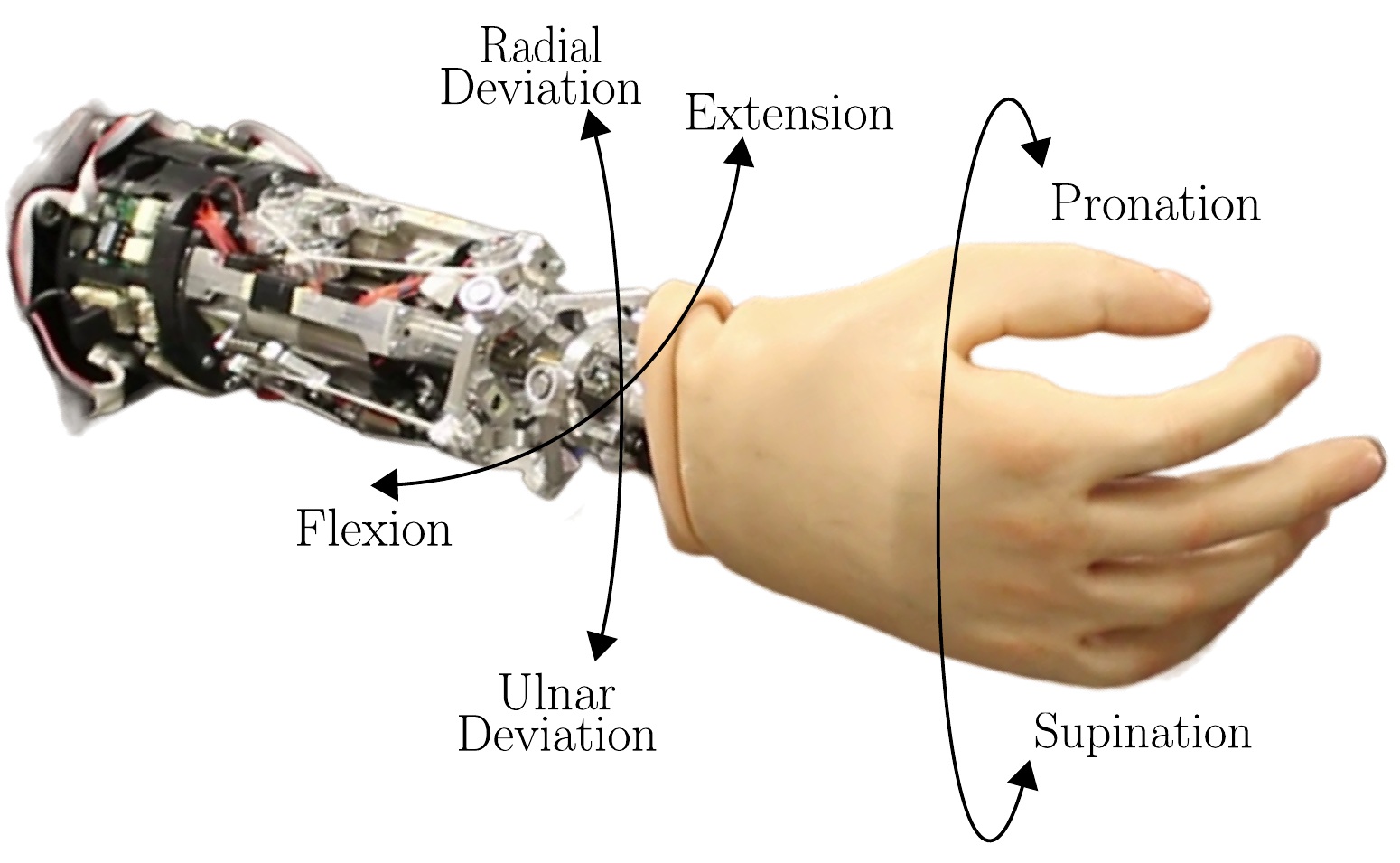}
    \caption{Representation of the device indicating its DoFs. In contrast to the human wrist, the Pronation/Supination motor solely rotates the end-effector rather than the entire forearm}
    \label{fig:General_Wrist}
\end{figure}

\section{Working Principles}
\label{sec:Working Principles}
The VS-Wrist, shown in Fig. \ref{fig:General_Wrist}, incorporates the parallel 2 DoFs VS joint in \cite{Lemerle2021Wrist} to achieve wrist flexion/extension (FE) and radial/ulnar deviation (RUD). Additionally, a serial motor unit is integrated to drive the pronation/supination (PS) of the wrist. The device weighs 1110 g, has a diameter of 70 mm, and is 170 mm long.
The parallel manipulator (PM), detailed in Fig.~\ref{fig:CADwrist}, ensures singularity-free motion of the coupler within a hemisphere. It features three legs evenly distributed around the base frame, each consisting of a serial arrangement of four non-coplanar revolute joints. The kinematics of the device draws inspiration from the Omni-Wrist III described in \cite{Rosheim2002OmniWrist}. 
However, our PM design introduces an additional motor unit and a non-linear elastic transmission to modulate the stiffness of the coupler. 
Leveraging the actuation redundancy, the stiffness regulation principle is achieved by reconfiguring the active length of the elastic elements through internal forces within the PM. 

The supplementary serial motor unit directly drives the axial rotation of the end-effector, transmitting this motion through two opposed universal joints.

\section{System Analysis}
\begin{figure}[b]
    \centering
    \includegraphics[width = 0.95\linewidth]{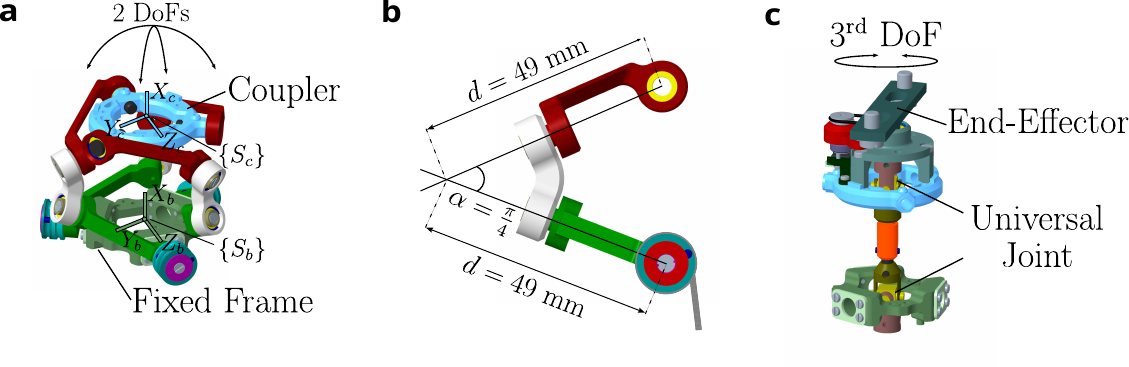}
    \caption{Schematic illustration depicting the kinematic configuration of the PM (\textbf{a}), a generic leg of the PM (\textbf{b}), and the PS transmission mechanism (\textbf{c})}
    \label{fig:CADwrist}
\end{figure}
\label{subsec:Kinematic Analysis}
To describe the posture of the PM, we define two reference coordinate frames as shown in Fig. \ref{fig:CADwrist}a: $\{S_b\} = \{O_b, X_b, Y_b, Z_b\}$, fixed to the base of the wrist, and $\{S_c\}~=~\{O_c, X_c, Y_c, Z_c\}$, attached to the coupler. 
The pose of the coupler with respect to (w.r.t.) $\{S_b\}$ can be described by the vector $\textbf{x} = {[} \alpha_x \enspace \alpha_y \enspace \alpha_z \enspace x_c \enspace  y_c \enspace z_c{]}^{\top}$, whose first three components represent Euler angles, while the last three defines the position of $O_c$. However, since the PM has 2 DoFs, the minimum parametrization ${u}~=~{[} \alpha_y \enspace \alpha_z{]}^{\top}$, representing the RUD and FE angles of the wrist, is sufficient to uniquely determine its pose.
To parametrize the homogeneous transformation from $\{S_b\}$ to $\{S_c\}$, we employ the Euler ZYX convention, resulting in
\begin{equation}
    \label{eq:pose_culer_cxtend}
    \text{T}_b^c (\textbf{x}) =
    \text{T}_{T_z}(z_c) \text{T}_{T_y}(y_c) \text{T}_{T_x}(x_c)  
    \text{T}_{R_z}(\alpha_z) \text{T}_{R_y}(\alpha_y) \text{T}_{R_x}(\alpha_x) \enspace , 
\end{equation}
where $\text{T}_{T_\ast}(p)$ and $\text{T}_{R_\ast}(p)$ are homogeneous transformation matrices that respectively encode a translation and a rotation along the generic $\ast$-axis of the quantity $p$.

\subsection{Forward Kinematics}
Fig. \ref{fig:CADwrist}b illustrates the kinematic chain of a generic leg of the PM. Given its joint configurations $q = {[}q_1 \enspace q_2 \enspace q_3 \enspace q_4{]}^{\top}$, the transformation matrix $\text{T}_b^c$ is obtained using the Denavit-Hartenberg parametrization, detailed in Table~\ref{tab:DHwrist}, as successive transformations between intermediate local frames yielding 
\begin{equation}
\label{eq:TBE_Q}
    \text{T}_b^c (q) = \text{T}_b^0(\eta) \text{T}_0^1(q_1) \text{T}_1^2(q_2) \text{T}_2^3(q_3) \text{T}_3^c(q_4) \enspace ,
\end{equation}
where $\text{T}_{i-1}^i = \text{T}_{R_z}(\theta_i) \text{T}_{T_z}(d_i) \text{T}_{R_x}(\alpha_i) \text{T}_{T_x}(a_i)$ defines the homogeneous transformation matrix between subsequent local frames and $\eta$ is the angular deviation of the leg w.r.t. $Z_b$. 

The PS rotation axis is always parallel to the $X_c$-axis. Therefore, indicating $\theta_{PS}$ as the related motor angle, the homogeneous transformation matrix encoding the pose of the end-effector becomes $\text{T}_b^{e}(q,\theta_{PS}) = \text{T}_b^c(q) \text{T}_{R_x}(\theta_{PS})$.

\begin{table}[t]
\begin{minipage}{0.45\linewidth}
\caption{Denavit Hartenberg parametrization of a generic leg of the PM. $d = 49$ mm, $\alpha = \frac{\pi}{4}$, and $\eta$ are fixed design parameters. The first rotation aligns the fixed base frame with the first joint axis of a given leg.}\label{tab:DHwrist}
\end{minipage}
\hfill
\begin{minipage}{0.48\linewidth}
\begin{tabular*}{0.92\linewidth}{@{\extracolsep{\fill}} c | c c c c}
\hline\noalign{\smallskip}
 $\mathbf{\{S_{i-1}\} \rightarrow \{S_{i}\}}$ & $\mathbf{\theta_i}$ & $\mathbf{d_i}$ & $\mathbf{a_i}$ & $\mathbf{\alpha_i}$\\
\noalign{\smallskip}\svhline\noalign{\smallskip}
$\{S_{b}\} \rightarrow \{S_{0}\}$ & 0 & 0 &  0 & $\eta$\\
$\{S_{0}\} \rightarrow \{S_{1}\}$ & $q_1$ & 0 & 0 & ${\pi}/{2}$\\
$\{S_{1}\} \rightarrow \{S_{2}\}$ & $q_2$ & $d$ & 0 & $-\alpha$\\
$\{S_{2}\} \rightarrow \{S_{3}\}$ & $q_3$ & $-d$ & 0 & ${\pi}/{2}$\\
$\{S_{3}\} \rightarrow \{S_c\}$ & $q_4$ & 0 &  0 & $\pi - \eta$\\
\noalign{\smallskip}\hline\noalign{\smallskip}
\end{tabular*}
\end{minipage}
\end{table}

\subsection{Inverse Kinematics}
Given the pose of the end-effector $\textbf{x}$, the Inverse Kinematics (IK) of the PM provides the joint angles of a chosen leg. 
By linearly combining the elements of $\text{T}_b^c$, the joint angles can be extracted as
\begin{equation}\label{eqn:IK_q1_sol}
\hspace{-0.45em}
\begin{cases}
    q_1 = \atantwo\left(
        x_c (1-c_{\alpha}) + (y_c c_{\eta}  + z_c s_{\eta})s_{\alpha} s_2 , \;
         d (1 - c_{\alpha}) (2 - (1 + c_{\alpha})c_2^2)
    \right) \\       
    q_2 = \acos\left(\frac{y_c s_\eta - z_c c_\eta}{d s_{\alpha}} \right) \hfill .\\
    q_3 = q_2 + \pi \\
    q_4 = -q_1
\end{cases}
\end{equation}
More detailed calculations can be found in \cite{Lemerle2021Wrist}.

\subsection{Differential Kinematics}
Differentiating \eqref{eqn:IK_q1_sol} results in a set of constraints on joint velocities dictated by the parallel architecture. Given that the PM has 2 DoFs, its posture $\textbf{x}$ can be expressed as a function of its minimum parametrization ${u}$ as detailed in \cite{Lemerle2021Wrist}.  Substituting $\textbf{x} = f({u})$ into \eqref{eqn:IK_q1_sol} and differentiating, we obtain
\begin{equation}
    \label{eq:diff_kin}
    \dot{Q} = \left[\left(\frac{\partial \mathcal{IK}_A(u)}{\partial u}\right)^\top \enspace \left(\frac{\partial \mathcal{IK}_B(u)}{\partial u}\right)^\top \enspace \left(\frac{\partial \mathcal{IK}_C(u)}{\partial u}\right)^\top\right]^\top \dot{u} = J_{IK} \dot{u} \;,
\end{equation}
where $Q = [q_A^\top \; q_B^\top \; q_C^\top]^\top$ represents the joint variables of every leg, $\mathcal{IK}_\ast(u)$ is the IK function of leg $\ast$ after substituting $\textbf{x} = f(u)$ in \eqref{eqn:IK_q1_sol}, and $J_{IK} \in \mathbb{R}^{12\times2}$ is the IK jacobian.

\subsection{Static Stiffness}\label{sec:Stiffness}
To achieve VS, the employed elastic transmission must be non-linear \cite{English1999VSPro}.
The non-linear elastic torque is approximated as $\tau_s = -2K\sinh{\left(\frac{\delta}{\delta_0}\right)}$, where $\delta$ is the deflection of the elastic mechanism, obtained as the difference between the first joint and motor angles (i.e., $\delta = q_1 - \theta$), and $K$ and $\delta_0$ are elastic parameters that shape the spring characteristics. The resulting stiffness function is $\sigma_s = 2\frac{k}{\delta_0}\cosh{\left(\frac{\delta}{\delta_0}\right)}$.
For simplicity, we model three identical springs approximating the elastic mechanism in \cite{Milazzo2023Wrist} with $K = 4$ $\frac{\text{Nmm}}{\text{rad}}$ and $\delta_0 = 0.32$ rad\textsuperscript{-1}.
 
For the kinetostatic duality, the actuated torques $\tau_a$ that balance an external wrench $w_u$ expressed in minimum parametrization hold $w_u = J_{a}^\top \tau_a$, where $J_{a} \in \mathbb{R}^{3\times2}$ is obtained by selecting only the first, fifth, and ninth rows of $J_{IK}$.
Since $\text{rank}(J_{a}^\top) = 2$, all actuated torques $\tau_a \in \text{ker}(J_{a}^\top)$ generate internal forces to the PM, thus they do not alter the equilibrium position of the coupler. 
To modulate these torques, a shift in the elastic element deflection is essential. Consequently, due to their non-linear characteristics, the stiffness of the elastic transmission must also change.

We define the static stiffness of the coupler $\Sigma_c \in \mathbb{R}^{2\times2}$ as 
\begin{equation}\label{eq:CouplerStiffness}
    \Sigma_c = \frac{\partial w_u}{\partial u} = \frac{\partial w_u}{\partial \tau_a} \frac{\partial \tau_a}{\partial q_a} \frac{\partial q_a}{\partial u} = J_a^\top \Sigma_s J_a
\end{equation}
where $\Sigma_s \in \mathbb{R}^{3\times3}$ is a diagonal matrix containing the stiffness $\sigma_s$ of the elastic transmission of each leg of the PM.
This formulation indicates that changing the spring preload affects the dimension of the stiffness ellipsoid, while its orientation essentially depends on posture, resembling human muscular impedance regulation \cite{Ajoudani2015Impedance}.

\subsection{Dynamic Model}
To simulate the behavior of the device, we implemented in Matlab Simulink the Euler-Lagrange equations describing the dynamics of the PM.
\begin{equation}\label{eq:dynamics}
    \begin{cases}
        B(Q)\ddot{Q} + C(Q,\dot{Q})\dot{Q} + D \dot{Q} + G(Q) + A^\top(Q)y = J^\top_w w_e + U_a^\top \tau_a\\
        \dot{Q} = J_{IK}(u)\dot{u}\\
        \ddot{Q} = J_{IK}(u) \ddot{u} + \dot{J}_{IK}(u,\dot{u})\dot{u}\\
    \end{cases}\enspace,
\end{equation}
where the last two equations account for the joint velocities and accelerations constraints given by the PM architecture, $B$ is the inertial matrix, $C$ is the Coriolis matrix, $D$ is the joint damping matrix, $G$ is the gravitational term, $A^\top$ is a base of the coupler constraints, $y$ parametrizes the reaction forces, $J_w$ is the geometric jacobian matrix of the PM, computed at the application point of the external wrench $w_e$, and $U_a$ is the underactuation matrix, considering that the actuated torques $\tau_a$ act directly only on the first joint of each leg.
We then project the equations into the null space of the parallel constraint forces by left-multiplying them with $J_{IK}^\top$, obtaining
\begin{equation}
\label{eq:Dyn_u}
J_{IK}^\top(BJ_{IK}\ddot{u} + (B\dot{J}_{IK} + CJ_{IK} + DJ_{IK})\dot{u} + G)= J_{IK}^\top(J_{w}^\top  w_{e} + U_a^\top \tau_{a}) \enspace,
\end{equation}
where functional dependency is omitted for clarity. 
We consider a point mass of 300 g, located 10 cm from the center of the coupler, to model an anthropomorphic end-effector. The inertial properties of every link of the PM are extracted from the CAD to obtain a reliable model. Motor dynamics are modeled with a low-pass system, while the non-linear elastic torque is shaped as described in Section~\ref{sec:Stiffness}. For simplicity, we assume linear viscous friction, even though it may vary with the preload.

\section{Control Strategy}
The IK solution is crucial to control wrist posture. Given a desired posture $u^r$, the solution $q_1^r$ from \eqref{eqn:IK_q1_sol} provides reference motor angles to achieve that posture.
As the PS unit directly actuates hand rotation and is decoupled from the PM kinematics, its motor reference coincides with the desired PS angle.

To regulate the stiffness of the coupler, we modulate internal torques through $\tau_a \in \text{ker}(J_{a}^\top)$, with $\text{dim}\left(\text{ker}(J_{a}^\top)\right) = 1$. 
Therefore, the actuated torques that regulate joint stiffness $\tau_\sigma$ can be expressed as $\tau_\sigma = \lambda N$, where $N \in \mathbb{R}^3$ forms a base of $\text{ker}(J_{a}^\top)$, and $\lambda \in \mathbb{R}$ parametrizes internal torques. 
Having a single DoF $\lambda$ to modulate joint stiffness limits reaching any generic configuration. To find $\lambda^r$ that achieves the closest reachable configuration to a given cartesian compliance reference $c^r$, we define an optimization problem as
\begin{equation}
    \label{eq:optimization}
    \lambda^r = \argmin_{\lambda \in \mathbb{R}} G(\lambda) \;, \text{where} \enspace G(\lambda) =  \norm{c - c^r}_F = \sqrt{\Tr\left((c - c^r)^\top(c - c^r)\right)} 
\end{equation}
where $\Tr(\ast)$ sums the diagonal elements and $\norm{\ast}_F$ represents the Frobenius norm of a given matrix $\ast$. 
We approach this optimization problem using the Levenberg-Marquardt method, imposing dynamics on the scalar parameter $\lambda^r$ as
\begin{equation}
    \label{eq:lambdadot}
    \dot{\lambda^r} = -\alpha \frac{H_g(\lambda)}{H_g^2(\lambda) + \nu}G(\lambda) \enspace,  
\end{equation}
where $H_g(\lambda) = \frac{\partial G(\lambda)}{\partial \lambda}$ and $\alpha, \nu \in \mathbb{R}$ act on the convergence rate and damping, respectively.

Therefore, the motor reference positions are given by 
\begin{equation}
    \theta_\ast^r = q_{1_\ast}^r + \delta_{0_\ast} \asinh \left(\frac{\lambda^r N_\ast}{2K_\ast}\right) \quad \text{for $\ast = A,\, B, \, C$} \enspace,
\end{equation}
where $N_\ast$ represents the leg $\ast$ component of $N$ and $\lambda^r$ is obtained by integrating \eqref{eq:lambdadot}.
Finally, a simple proportional controller regulates the motor positions.

\section{Simulation Results}
\begin{figure}[b]
\sidecaption[b]
    \centering
    \includegraphics[width = 7 cm]{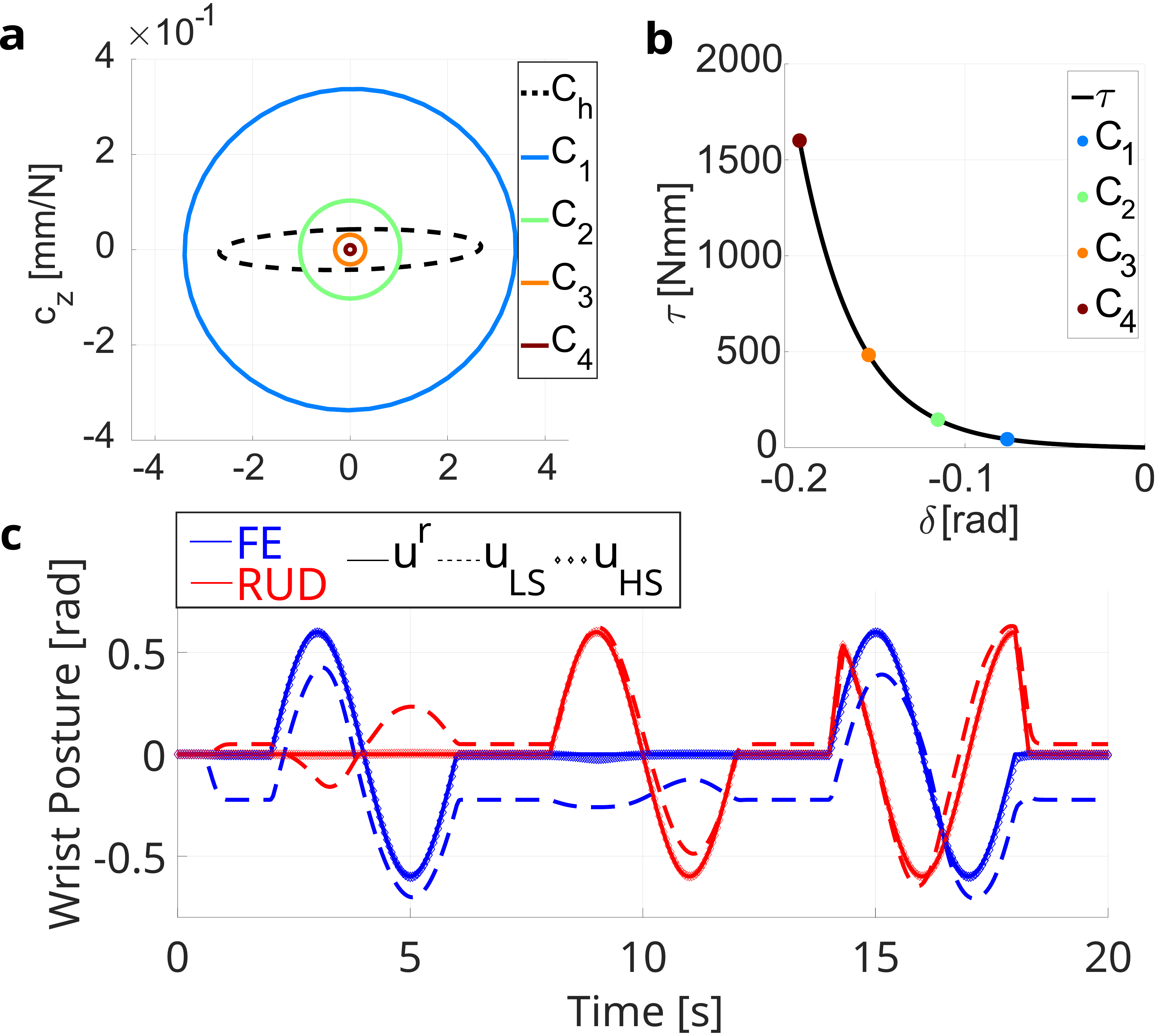}
    \caption{Simulation results illustrating the wrist stiffness and motion behavior. \textbf{a} Cartesian compliance of the wrist in the central position across four distinct stiffness levels, juxtaposed with human wrist compliance $C_h$ extracted from \cite{Pando2014Wrist}. \textbf{b} Corresponding elastic torque from the actuators. \textbf{c} Wrist motion behavior under loaded conditions, emphasizing the differences between the soft (LS, dashed lines) and rigid (HS, diamond lines) configurations. Continuous lines denote posture references. Blue lines indicate the FE angles, while red lines represent the RUD angles} 
    \label{fig:Exp_ARK}
\end{figure}
Fig. \ref{fig:Exp_ARK} summarizes key simulation results. Fig. \ref{fig:Exp_ARK}a and \ref{fig:Exp_ARK}b illustrate the impact of modulating the parameter $\lambda$ on joint stiffness and elastic torque, respectively.
Due to the pronounced non-linearity of the elastic mechanism, a moderate torque range yields a significant stiffness spectrum.
Fig. \ref{fig:Exp_ARK}c depicts the wrist motion at low (LS, dashed lines) and high (HS, diamond lines) stiffness configurations.
A load of 1.5 kg is applied at the center of the hand at 0.5 s. As anticipated, the rigid configuration effectively resists external perturbation, whereas the soft configuration is markedly influenced by the external load.
This observation is corroborated by the Root Mean Square (RMS) values of the posture error $e$ and elastic torque in the two trials. Specifically, we find $RMS(e^L) = 0.15$ rad, $RMS(e^H) = 0.007$ rad, $RMS(\tau_a^L) = 0.07$ Nm, and $RMS(\tau_a^H) = 1.72$ Nm.

\section{Conclusions}
This study details the theoretical model of a VS 3 DoFs wrist with redundant elastic actuation. Leveraging a parallel architecture, the device achieves stiffness modulation with a compact design. Simulation outcomes validated the effectiveness of the proposed control strategy in independently regulating joint position and overall stiffness, thus allowing the device to adapt to various environmental situations or tasks. 
Given its morphological and functional resemblance to the human wrist, the device holds promise for applications in prosthetics and humanoid robotics.

\begin{acknowledgement}
This research has received funding from the European Union’s ERC program under the Grant Agreement No. 810346 (Natural Bionics).
\end{acknowledgement}


\begin{thebibliography}{99.}%

\bibitem{Phan2020Tooling}  Phan G-H, Hansen C, Tommasino P, Budhota A, Mohan DM, Hussain A, Burdet E, Campolo D (2020) Estimating Human Wrist Stiffness during a Tooling Task. Sensors, 20(11): 3260

\bibitem{Osu2004Optimal} Osu R, Kamimura N, Iwasaki H et al (2004). Optimal impedance control for task achievement in the presence of signal-dependent noise. Journal of Neurophysiology 92(2): 1199-1215

\bibitem{Borzelli2018Muscles} Borzelli D, Cesqui B, Berger DJ et al (2018). Muscle patterns underlying voluntary modulation of co-contraction. PLoS One 13(10)

\bibitem{Wolf2016VSA} Wolf S, Grioli G, Eiberger O et al (2015). Variable stiffness actuators: Review on design and components. IEEE/ASME transactions on mechatronics 21(5): 2418-2430

\bibitem{Grioli2015User} Grioli G, Wolf S, Garabini M et al (2015). Variable stiffness actuators: The user’s point of view. The International Journal of Robotics Research 34(6): 727-743

\bibitem{Lemerle2021Wrist} Lemerle S, Catalano MG, Bicchi A et al (2021). A configurable architecture for two degree-of-freedom variable stiffness actuators to match the compliant behavior of human joints. Frontiers in Robotics and AI. \url{ https://doi.org/10.3389/frobt.2021.614145}

\bibitem{Milazzo2023Wrist} Milazzo G, Catalano MG, Bicchi A, Grioli G (2023). Modeling and Control of a Novel Variable Stiffness three DoFs Wrist. Preprint at \url{https://arxiv.org/abs/2305.16154}

\bibitem{Rosheim2002OmniWrist} Rosheim ME, Sauter GF (2002). New high-angulation omnidirectional sensor mount. In Free-space laser communication and laser imaging II. \url{https://doi.org/10.1117/12.465912}

\bibitem{English1999VSPro} English CE, Russell D (1999). Mechanics and stiffness limitations of a variable stiffness actuator for use in prosthetic limbs. \url{https://doi.org/10.1016/S0094-114X(98)00026-3}

\bibitem{Ajoudani2015Impedance}  Ajoudani A, Fang C, Tsagarakis NG et al (2015). A reduced-complexity description of arm endpoint stiffness with applications to teleimpedance control. \url{https://doi.org/10.1109/IROS.2015.7353495}

\bibitem{Pando2014Wrist} Pando AL, Lee H, Drake WB et al. (2014). Position-dependent characterization of passive wrist stiffness. IEEE Transactions on Biomedical Engineering. 61(8): 2235-2244.
\end{thebibliography}
\end{document}